\documentclass[11pt]{article}

\usepackage[final]{acl}

\usepackage{times}
\usepackage{latexsym}

\usepackage[T1]{fontenc}

\usepackage[utf8]{inputenc}

\usepackage{microtype}

\usepackage{inconsolata}

\usepackage{graphicx}
\usepackage{amsmath}
\usepackage{booktabs}
\usepackage{multirow}
\usepackage{xcolor}
\usepackage{algorithm}
\usepackage{algorithmic}
\usepackage{hyperref}
\usepackage[normalem]{ulem}

\usepackage[most]{tcolorbox}
\tcbuselibrary{listings,breakable}

\newtcblisting{promptbox}{
  enhanced,
  breakable,
  listing only,
  colback=gray!4,
  colframe=gray!45,
  boxrule=0.4pt,
  arc=1mm,
  left=1mm,
  right=1mm,
  top=0.8mm,
  bottom=0.8mm,
  listing options={
    basicstyle=\ttfamily\scriptsize,
    breaklines=true,
    breakatwhitespace=false,
    columns=fullflexible,
    keepspaces=true,
    showstringspaces=false,
    tabsize=2
  }
}

\title{\textsc{SkillAlign}: Aligning Skill Interfaces for LLM-based Agents}

\author{
 \textbf{Shuo Ren\textsuperscript{1}},
 \textbf{Xiaomian Kang\textsuperscript{1}},
 \textbf{Jiajun Zhang\textsuperscript{1,2,3}\thanks{Corresponding author.}}
\\
 \textsuperscript{1}Institute of Automation, Chinese Academy of Sciences\\
 \textsuperscript{2}School of Artificial Intelligence, University of Chinese Academy of Sciences\\
 \textsuperscript{3}Wuhan AI Research
\\
\texttt{shuo.ren@ia.ac.cn}, \texttt{\{xiaomian.kang,jjzhang\}@nlpr.ia.ac.cn}
}

\begin{document}
\maketitle
\begin{abstract}
Language-model agents increasingly rely on skills: reusable procedural knowledge for reasoning, tool use, and interaction. Existing work studies how skills are acquired, retrieved, compressed, or composed, but often assumes that once a skill is selected, its interface to the agent is fixed. We argue that this overlooks a key source of skill utility: the same skill can help, distract, or mislead depending on how it is exposed. We propose \textsc{SkillAlign}, a provider-agnostic framework that represents candidate skills as multi-view procedural cards and renders them through alternative exposure interfaces, including full instructions, hints, compressed summaries, workflows, or no exposure. This enables counterfactual evaluation where the task, agent, and candidate skills are fixed while only the exposure interface varies. Across ALFWorld and SkillsBench, we show that exposure form substantially affects task success and rendered context cost, and that compact top-\(k\) exposure can outperform full-library injection. We further conduct a replay-based policy-learning analysis on ALFWorld, showing that adaptive exposure contains learnable signal but remains far from oracle selection. Our results suggest that skill-augmented agents should optimize not only which skills to use, but also how those skills are presented.
\end{abstract}

\section{Introduction}

Large language model agents increasingly rely on \textbf{skills}: reusable procedural knowledge that can guide agents through recurring patterns of interaction, reasoning, and tool use. Unlike ordinary tool APIs, which expose external capabilities, skills often encode higher-level behavioral strategies: how to debug a software failure, navigate a household environment, compare candidate products, collect evidence for a claim, or follow domain-specific conventions. Recent surveys characterize skills as procedural artifacts that coordinate tools, memory, and runtime context under task-specific constraints, bridging the ``procedural gap'' between raw tool access and robust task execution \citep{xu2026agentskills,jiang2026sokskills,zhou2026comprehensivesurvey}. In this sense, skills act as the operational layer of agent systems: they externalize know-how that would otherwise need to be rediscovered from scratch in every task.

The skill-agent ecosystem has expanded rapidly across the full skill lifecycle. Skills can be represented as text-based instructions, code-backed routines, or hybrid artifacts \citep{shinn2023reflexion,zhao2024expel,wang2024awm,wang2023voyager,wang2024codeact,zheng2025skillweaver}; acquired from execution traces, task requirements, or external corpora \citep{wang2023voyager,ni2026trace2skill,qian2023creator,qin2024toolllm}; retrieved and selected through dense, generative, or structure-aware mechanisms \citep{wang2025toolgen,wang2026graphskill,zheng2026skillrouter,liu2026gos,li2026organizing}; and further compressed, compiled, structured, or evolved as environments change \citep{gao2026skillreducer,chen2026skvm,xia2026grasp,lu2026contractskill,zhang2026skillflow}. This lifecycle view shows that skills are no longer merely prompt snippets or isolated tools, but persistent procedural objects embedded in a broader ecosystem of representation, acquisition, retrieval, execution, and evolution \citep{zhou2026comprehensivesurvey}.

However, this lifecycle also reveals a missing dimension. Most work asks how skills should be written, acquired, retrieved, compressed, composed, or updated. These are essential questions, but they do not fully determine how a retrieved skill should affect the agent at inference time. A skill is not automatically useful simply because it is relevant. It may be too long, too vague, too specific, too outdated, too weakly grounded in the current environment, or too high-level for the agent’s current state. Conversely, a skill that fails as a full document may still help as a short reminder, a compact affordance-and-risk summary, a workflow template, or a compact operational summary. We therefore argue that skills should be viewed as externalized procedural priors whose effect depends not only on their content, but also on the interface through which they condition the agent. The practical question is not only which skill should be selected, but also how the selected skill should be exposed.

In this paper, we propose \textsc{SkillAlign}, a framework for skill-interface alignment in LLM agents. SkillAlign separates \emph{skill selection} from \emph{skill exposure}: upstream providers determine which skills are available, while SkillAlign determines how those skills condition the agent. Given candidate skills from any provider, SkillAlign converts each raw skill into a multi-view procedural card and renders the candidate set through an exposure interface, such as a full document, short hint, compressed operational summary, workflow, or no exposure. This turns exposure into a controlled decision variable, allowing us to evaluate whether different interfaces for the same task and candidate skills lead to different trajectories, costs, and failures. SkillAlign further supports learning an exposure policy from counterfactual outcomes, providing a path toward adaptive interface selection.

We evaluate SkillAlign on ALFWorld and SkillsBench under multiple skill providers and agent backends. Our experiments compare fixed exposure modes and learned exposure policies while holding candidate skills fixed when needed. The results show that exposure form changes task success and context cost even when the same skills are selected. In particular, concise exposure can match or approach full-document injection in some settings, while learned selectors reveal substantial headroom over any single fixed interface.

Our contributions are as follows:
\begin{itemize}
    \item We identify \emph{skill-interface alignment} as a missing problem in skill-augmented agents: skill utility depends not only on skill relevance or quality, but also on how the skill is exposed to the agent.

    \item We propose \textsc{SkillAlign}, an exposure framework that represents skills as multi-view procedural cards and renders candidate skills through alternative interfaces, including full instructions, hints, compressed summaries, workflows, or no exposure.

    \item We introduce a counterfactual evaluation protocol that holds the task, agent backend, and candidate skills fixed while varying only the exposure interface, enabling direct measurement of interface-induced changes in task success and rendered skill-context cost.

    \item We evaluate SkillAlign across ALFWorld and SkillsBench with multiple providers and agent backends, showing that exposure form is a first-class factor in skill-augmented performance and that adaptive exposure contains learnable signal.
\end{itemize}

\section{Related Work}

\subsection{Agent Skills as Procedural Priors}

Agent skills externalize reusable procedural knowledge for LLM agents. Recent surveys define skills as modular artifacts that coordinate tools, memory, and runtime context under task-specific constraints, thereby bridging the procedural gap between raw tool access and reliable execution \citep{xu2026agentskills,jiang2026sokskills,zhou2026comprehensivesurvey}. This view distinguishes skills from ordinary tools: tools expose operations, while skills package know-how for when and how those operations should be used in context. Skills therefore act as procedural priors that bias the agent toward reusable patterns of reasoning, interaction, and tool use.

Existing systems instantiate such priors in different forms. Text-based skills encode reflections, lessons, reasoning templates, or workflow memories; code-backed skills include executable scripts, API calls, or programmatic routines; and hybrid skills combine textual guidance with executable or structured runtime support \citep{shinn2023reflexion,zhao2024expel,yang2024bot,wang2024awm,ni2026trace2skill,wang2023voyager,wang2024codeact,yang2024sweagent,zhang2023toolcoder,zheng2025skillweaver}. SkillAlign builds on this procedural-prior view but studies a different design variable: not how a skill is acquired or represented, but how an already available skill should condition the agent in a given task state.

\begin{figure*}[tp]
    \centering
    \includegraphics[width=0.9\linewidth, height=160pt]{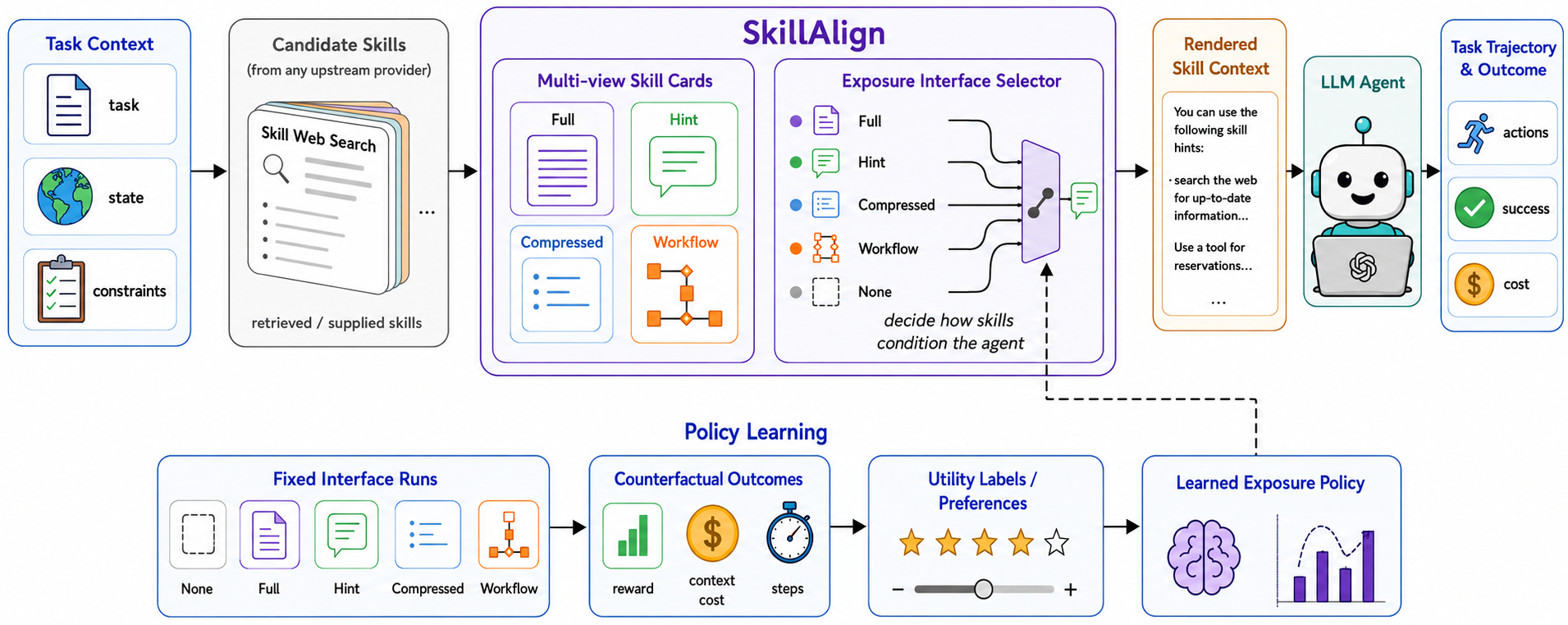}
    \caption{
    Overview of SkillAlign. Given a task context and candidate skills from any upstream provider, SkillAlign constructs multi-view skill cards and selects an exposure interface that determines how the skills condition the agent. Fixed-interface runs provide counterfactual outcomes for evaluating exposure effects. These outcomes can also supervise an exposure policy that selects interfaces adaptively.
    }
    \label{fig:skillalign_overview}
\end{figure*}

\subsection{Skill Acquisition, Retrieval, and Structured Use}

A large body of work studies how skills are obtained and made available. Skills may be distilled from prior trajectories, generated from current task requirements, extracted from external corpora, or accumulated through lifelong evolution \citep{wang2023voyager,shinn2023reflexion,zhao2024expel,wang2024awm,ni2026trace2skill,qian2023creator,qin2024toolllm,zhang2026memskill,xia2026skillrl,yang2026autoskill,wang2025reinforcement}. These works address where skills come from and how they are improved; SkillAlign instead starts after candidate skills are available.

Another line of work studies retrieval, routing, and selection from growing skill libraries \citep{wang2023voyager,wang2025toolgen,wang2026graphskill,zheng2026skillrouter,liu2026gos,li2026organizing}. Dense and sparse retrieval match task contexts to skill descriptions or metadata, generative retrieval predicts tool or skill identifiers, and structure-aware methods exploit hierarchies, dependencies, or graph relations. Other systems optimize the skill artifact itself \citep{gao2026skillreducer,chen2026skvm,wang2026webxskill,lu2026contractskill,xia2026grasp}: SkillReducer compresses verbose skills, SkVM compiles skills across model-harness pairs, WebXSkill bridges textual workflow guidance with executable web programs, ContractSkill formalizes executable contracts, and GraSP composes retrieved skills into verified graph-structured procedures . These directions mainly ask which skills should be surfaced, shortened, compiled, or composed. SkillAlign is complementary: it asks how selected skills should be exposed to the agent.

\section{Method}
\label{sec:method}

SkillAlign focuses on a design choice that is usually implicit in skill-augmented agents: given a set of candidate skills, \emph{through what interface should these skills condition the agent}? As illustrated in Figure~\ref{fig:skillalign_overview}, SkillAlign takes the task context, agent context, and candidate skills as input. The candidate skills may come from any upstream source, such as exhaustive skill provision, vector retrieval, graph-based retrieval, or oracle annotations. SkillAlign then represents each raw skill as a multi-view procedural card and renders the candidate set through a selected exposure interface, such as a full instruction, short hint, compressed summary, workflow, or no exposure. The rendered skill context is injected into the agent's ordinary interaction loop, and the resulting trajectory is evaluated by the benchmark. Our main analysis uses fixed exposure modes to create counterfactual trajectories, while learned exposure policies use these outcomes to study adaptive interface selection. \textbf{Note that}, in this work, the learned policy is intentionally lightweight: it is used to validate the feasibility of adaptive exposure rather than to fully solve interface selection. We therefore report it as a policy-learning analysis and compare it against oracle exposure to quantify the remaining headroom.

\subsection{Skill Exposure as an Interface Decision}
\label{sec:skill_exposure}

Let \(x\) denote the task context, \(a\) the agent context, and \(S_x=\{s_1,\ldots,s_k\}\) the candidate skills available for the task. SkillAlign treats \(S_x\) as given: the candidate set may come from exhaustive skill provision, vector retrieval, graph-based retrieval, or oracle annotations, but retrieval itself is not our object of optimization. Instead, SkillAlign chooses an exposure interface \(m\):

\[
m = \pi(x,a,S_x), \qquad m \in \mathcal{M},
\]

and renders the same candidate skills into an agent-facing context:

\[
r = \mathrm{Render}(S_x,m;x,a).
\]

The rendered context \(r\) does not replace or modify the original skill artifact. It is the interface through which the skill conditions the agent at inference time. In our main experiments,

\[
\mathcal{M}=\{
\texttt{none},
\texttt{full},
\texttt{hint},
\texttt{compressed},
\texttt{workflow}
\}.
\]

These interfaces define different strengths and structures of conditioning. \texttt{full} exposes the original skill document; \texttt{hint} gives a concise task-facing cue; \texttt{compressed} preserves key affordances, assumptions, constraints, and risks; \texttt{workflow} emphasizes ordered procedural guidance; and \texttt{none} suppresses skill text when candidate skills are unnecessary or potentially harmful. This formulation makes exposure a controlled variable: for the same task, agent, and candidate skills, different interfaces may induce different trajectories.

\subsection{Multi-View Skill Cards and Rendering}
\label{sec:skill_cards}

Raw skills are heterogeneous: they may be natural-language instructions, reflections, workflow memories, scripts, API recipes, repository-specific procedures, or hybrid packages. SkillAlign represents each skill as a \emph{multi-view skill card}, preserving the original artifact while exposing alternative task-facing views (see Appendix \ref{app:skill_cards}):

\[
c(s)=\{v_{\mathrm{full}},v_{\mathrm{hint}},v_{\mathrm{comp}},v_{\mathrm{flow}},\eta\},
\]

where \(v_{\mathrm{full}}\) is the original skill content, \(v_{\mathrm{hint}}\) is a concise cue, \(v_{\mathrm{comp}}\) is a compact affordance-and-risk view, \(v_{\mathrm{flow}}\) is an ordered workflow view, and \(\eta\) stores metadata such as skill id, provenance, source path, and retrieval-facing summary.

The views differ semantically rather than merely in length. The full view preserves execution-critical details such as assumptions, examples, scripts, and edge cases. The hint view acts as a weak activation signal. The compressed view summarizes applicability, actions, expected outputs, and mismatch risks. The workflow view emphasizes ordering and checkpoints. Given an interface \(m\), the renderer selects the corresponding view and constructs the injected context, framing skills as supportive priors rather than instructions that override task constraints or environment feedback.

\subsection{Fixed Exposure Modes and Counterfactual Evaluation}
\label{sec:fixed_modes}

The main evidence for SkillAlign comes from fixed exposure modes. A fixed policy always chooses the same interface,

\[
\pi_m(x,a,S_x)=m,
\]

and for each task and candidate source, we run the agent under multiple interfaces over the same candidate skill set:

\[
\{\tau_x^m: m\in\mathcal{M}\}.
\]

These counterfactual trajectories isolate the effect of exposure form. If the task, agent backend, and candidate skills are held fixed, but performance changes across \texttt{full}, \texttt{hint}, \texttt{compressed}, \texttt{workflow}, and \texttt{none}, then the interface is not a cosmetic formatting choice but a substantive factor in skill utility.

We use these trajectories to measure task success or reward and injected context cost. A task is counted as a negative-transfer case for interface \(m\) if the no-exposure run succeeds but the run with \(m\) fails, or if scalar reward decreases under \(m\). Repeating the same comparison under different candidate sources allows us to analyze how exposure interacts with upstream skill provision without making retrieval itself the contribution.

\subsection{Exploratory Exposure Policy Learning}
\label{sec:policy_learning}

We further study whether exposure choice can be learned from counterfactual outcomes. This component is not required by the SkillAlign framework, but it provides a way to estimate how much interface selection can be improved beyond fixed exposure modes.

For each task and candidate set, fixed-mode runs provide outcomes under all interfaces from an auxiliary dataset. We convert them into a utility score:

\[
U_x(m)=R_x(m)-\alpha C_x(m)-\beta T_x(m),
\]

where \(R_x(m)\) is reward or success, \(C_x(m)\) is normalized context cost, and \(T_x(m)\) is normalized trajectory length. The best-utility interface gives an SFT label,

\[
m_x^*=\arg\max_{m\in\mathcal{M}} U_x(m),
\]

while interface pairs with sufficient utility margin form preference examples for DPO. The learned policy receives the task description, candidate skill-card content, and source metadata, then predicts a single interface label for the candidates.

In our experiments, we train lightweight LoRA adapters on Qwen3-4B model \cite{yang2025qwen3}. SFT serves as a stable first attempt at learning exposure decisions, while DPO tests whether preference optimization over counterfactual utilities can further improve interface choice. Note that the current exposure policy learning is deliberately basic and lightweight, intended to verify the feasibility of optimizing skill interfaces rather than to solve the problem fully. More details are in Appendix \ref{app:policy_training}.

\section{Experiments}
\label{sec:experiments}

We evaluate whether skill exposure is a first-class factor in skill-augmented agents. The central question is: given the same task, agent backend, and candidate skill set, does changing the exposure interface alter task success or context cost? To answer this question, we conduct controlled counterfactual evaluations across interactive and tool-use agent benchmarks. For each candidate source, we render the same skills through different interfaces---\texttt{full}, \texttt{hint}, \texttt{compressed}, \texttt{workflow}, or \texttt{none}---and compare the resulting trajectories. This design directly tests our hypothesis that skills act as externalized procedural priors whose effect depends on how they condition the agent.

\subsection{Setup}
\label{sec:exp_setup}

\paragraph{Benchmarks.}
We evaluate on two benchmarks for skill-augmented agents. \textbf{ALFWorld} \citep{shridhar2021alfworld} is a text-based embodied environment where agents complete multi-step household tasks by observing textual states and issuing textual actions. Following ReAct-style evaluation \citep{yao2023react}, we use the text-only setting with 140 evaluation episodes. \textbf{SkillsBench} \citep{li2026skillsbench} contains real-world technical tasks paired with curated skills, covering domains such as data analysis, scientific computing, financial modeling, and engineering workflows. We evaluate on 94 SkillsBench task instances using Harbor and \texttt{Terminus-2} \citep{Harbor_Framework} execution harness. We use \textbf{ScienceWorld} \citep{wang2022scienceworld} for auxiliary exposure-policy data generation.

\paragraph{Skill libraries.}
We denote a skill library as \texttt{skills\_N}, where \(N\) is the number of available skills, not the number of evaluation tasks. The main experiments use \texttt{skills\_1000}. For scaling analysis, we additionally evaluate \texttt{skills\_200}, \texttt{skills\_500}, and \texttt{skills\_2000}. Skill cards are constructed from raw skill artifacts before evaluation. For overlapping skill ids with identical raw content, cards are reused across libraries.

\paragraph{Candidate sources.}
SkillAlign is provider-agnostic: it operates on candidate skills supplied by upstream sources. The main experiments use three generic candidate sources: \texttt{none}, which injects no skill context; \texttt{all}, which makes the available skill library available without task-specific filtering; and \texttt{vector-top\(k\)}, which retrieves skills using semantic similarity via \texttt{text-embedding-3-large} embeddings. We use \(k=5\) for ALFWorld and \(k=8\) for SkillsBench. These sources represent no exposure, broad skill availability, and standard top-\(k\) retrieval. We additionally evaluate a graph-based provider in Section~\ref{sec:provider_influence} to test whether exposure effects persist under structured skill selection.

\paragraph{Exposure interfaces.}
For each candidate source, we compare five exposure interfaces: \texttt{none}, \texttt{full}, \texttt{hint}, \texttt{compressed}, and \texttt{workflow}. The \texttt{full} interface injects the original skill content; \texttt{hint} provides concise task-facing cues; \texttt{compressed} summarizes affordances, assumptions, constraints, and risks; \texttt{workflow} presents ordered procedural guidance; and \texttt{none} suppresses skill text.

\paragraph{Agent backends and metrics.}
We evaluate three backends: \texttt{minimax-m2.7}, \texttt{gpt-5.3-codex}, and \texttt{glm-5}, accessed through API services. We report success rate (SR) or average reward (AR), together with rendered skill-context cost \(C\). For ALFWorld, reward is binary and equals success rate. \(C\) denotes the token length of the rendered skill context, reported in thousands of tokens. For overlong all-skill settings, \(C\) reflects the rendered exposure scale before backend-specific truncation or rejection.

\subsection{Main Results}
\label{sec:main_results}

The main comparison asks whether exposure form changes agent behavior when the candidate skills are held fixed. Table~\ref{tab:main_results} reports both performance and rendered skill-context cost under \texttt{skills\_1000} across three agent backends. Rows specify the candidate source and exposure interface, while columns report performance--cost pairs for each backend. 

\begin{table*}[t]
\centering
\scriptsize
\resizebox{\linewidth}{!}{
\begin{tabular}{l|cc|cc|cc|cc|cc|cc}
\toprule
\multirow{3}{*}{\textbf{Configuration}}
& \multicolumn{6}{c|}{\textbf{ALFWorld}} 
& \multicolumn{6}{c}{\textbf{SkillsBench}} \\
\cmidrule(lr){2-7} \cmidrule(lr){8-13}
& \multicolumn{2}{c|}{\textbf{MiniMax}}
& \multicolumn{2}{c|}{\textbf{GPT-Codex}}
& \multicolumn{2}{c|}{\textbf{GLM}}
& \multicolumn{2}{c|}{\textbf{MiniMax}}
& \multicolumn{2}{c|}{\textbf{GPT-Codex}}
& \multicolumn{2}{c}{\textbf{GLM}} \\
& SR $\uparrow$ & C $\downarrow$
& SR $\uparrow$ & C $\downarrow$
& SR $\uparrow$ & C $\downarrow$
& AR $\uparrow$ & C $\downarrow$
& AR $\uparrow$ & C $\downarrow$
& AR $\uparrow$ & C $\downarrow$ \\
\midrule
No Skill / None
& 50.7 & 0.0 & 89.3 & 0.0 & 73.6 & 0.0
& 17.2 & 0.0 & 27.4 & 0.0 & 22.6 & 0.0 \\
\midrule
All + Full
& 47.9 & 900.0 & 88.6 & 900.0 & 70.7 & 900.0
& 14.5 & 1080.0 & 24.8 & 1080.0 & 20.4 & 1080.0 \\
All + Hint
& 68.6 & 6.0 & 92.1 & 6.0 & 80.7 & 6.0
& 16.8 & 7.5 & 29.4 & 7.5 & 24.0 & 7.5 \\
All + Compressed
& 72.1 & 25.2 & 93.6 & 25.2 & 82.1 & 25.2
& \underline{17.2} & 31.0 & 30.8 & 31.0 & \underline{25.1} & 31.0 \\
All + Workflow
& 61.4 & 282.5 & 90.0 & 282.5 & 76.4 & 282.5
& 15.6 & 335.0 & 27.6 & 335.0 & 22.8 & 335.0 \\
\midrule
Vector-top\(k\) + Full
& \underline{73.6} & 4.7 & 92.9 & 4.7 & 80.7 & 4.7
& 16.4 & 9.0 & \underline{31.8} & 9.0 & 24.8 & 9.0 \\
Vector-top\(k\) + Hint
& \textbf{75.0} & 0.08 & \textbf{94.3} & 0.08 & \textbf{83.6} & 0.08
& 15.6 & 0.13 & 28.9 & 0.13 & 23.4 & 0.13 \\
Vector-top\(k\) + Compressed
& 69.3 & 0.30 & \underline{94.0} & 0.30 & \underline{82.9} & 0.30
& \textbf{18.1} & 0.55 & \textbf{32.2} & 0.55 & \textbf{26.2} & 0.55 \\
Vector-top\(k\) + Workflow
& 64.3 & 1.5 & 91.4 & 1.5 & 78.6 & 1.5
& 16.2 & 2.9 & 29.6 & 2.9 & 24.4 & 2.9 \\
\bottomrule
\end{tabular}
}
\caption{
Main SkillAlign results under \texttt{skills\_1000}. ALFWorld reports success rate (SR), and SkillsBench reports average reward (AR). \(C\) denotes the token length of the rendered skill context, reported in thousands of tokens. Each row fixes a candidate source and exposure interface; within each candidate-source block, candidate skills are fixed and only the exposure interface changes. Best performance results are bolded and second-best results are underlined within each benchmark--backend block.  Task-level bootstrap 95\% confidence intervals are reported in Appendix~\ref{app:uncertainty}. Lower cost is better.
}
\label{tab:main_results}
\end{table*}

\paragraph{Exposure is a first-class conditioning variable.}
Across benchmarks and backends, changing only the exposure interface leads to substantial performance differences. Under the same \texttt{all} candidate source, ALFWorld success varies from 47.9 to 72.1 for MiniMax and from 70.7 to 82.1 for GLM. Under \texttt{vector-top\(k\)}, the same candidate set also yields different outcomes across interfaces, such as 64.3--75.0 on ALFWorld with MiniMax and 28.9--32.2 on SkillsBench with GPT-Codex. This supports our claim that a skill is not simply useful or harmful in isolation; it becomes useful through an interface that controls how strongly and in what form it conditions the agent.


\paragraph{Candidate quality and exposure form interact.}
Although SkillAlign does not propose a new retriever, the results show that exposure effects depend on the quality of the candidate set. \texttt{All + Compressed} improves over \texttt{All + Full}, but it still aggregates compressed information from all 1000 skills and can contain many irrelevant procedural priors. In contrast, \texttt{Vector-top\(k\)} substantially reduces the candidate set before exposure. This pattern shows that candidate filtering and exposure design are complementary: retrieval reduces irrelevant skill mass, while SkillAlign controls how the remaining procedural priors are presented.
\begin{figure}[t]
    \centering
    \includegraphics[width=\linewidth]{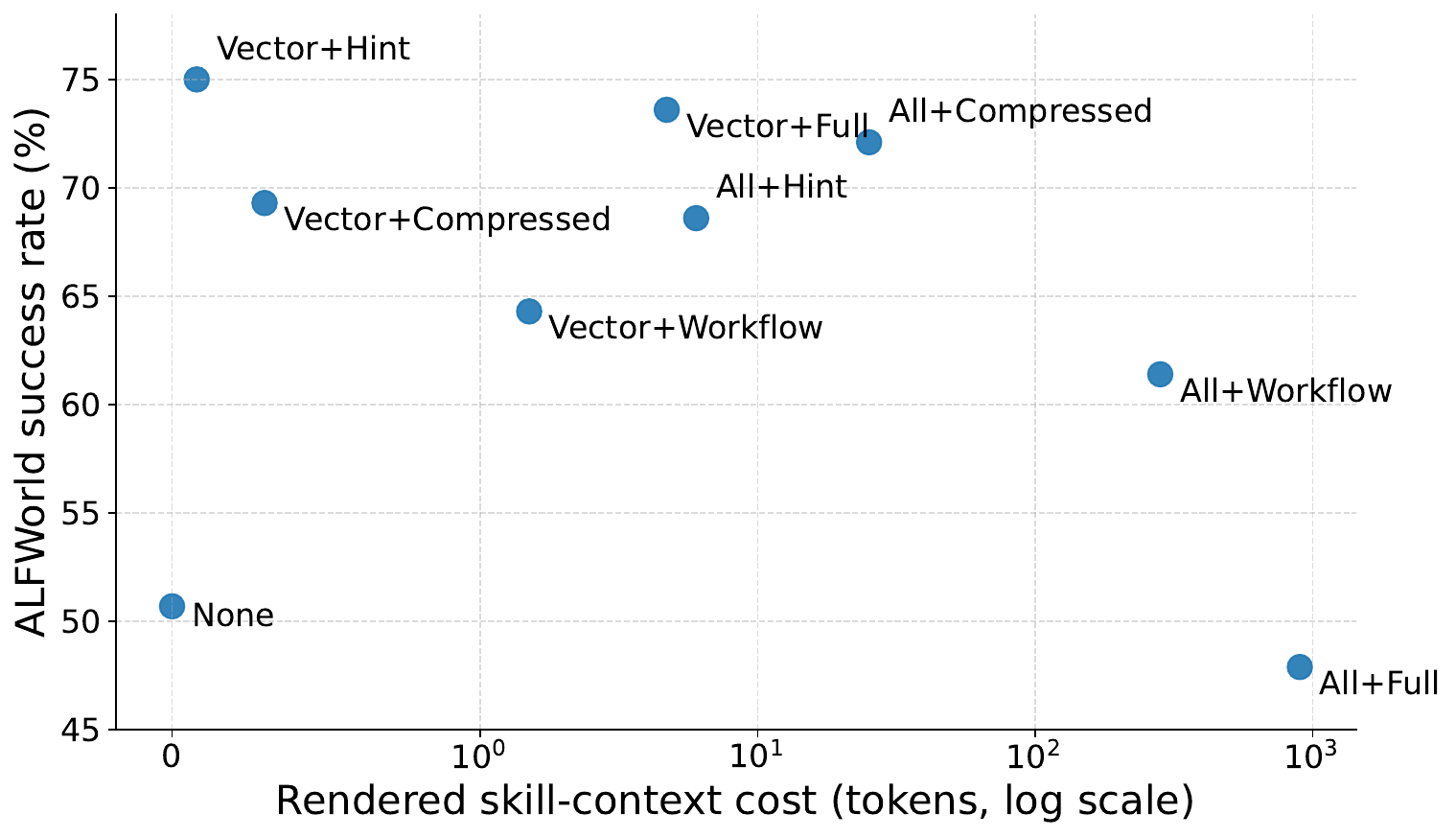}
    \caption{
    Cost--performance tradeoff of exposure interfaces under MiniMax M2.7 on ALFWorld. Each point corresponds to a row in the ALFWorld--MiniMax block of Table~\ref{tab:main_results}. Interfaces closer to the upper-left provide better utility per unit of rendered skill-context cost.
    }
    \label{fig:cost_tradeoff}
\end{figure}
\paragraph{Compact exposure is often more efficient.}
Compact interfaces often achieve strong reward at much lower cost. For example, \texttt{Vector-top\(k\) + Hint} and \texttt{Vector-top\(k\) + Compressed} achieve top-ranking results in multiple columns while remaining below 1K token-equivalent units for compact views. Among them, \texttt{hint} is the shortest interface, while \texttt{compressed} uses more context to preserve triggers, state assumptions, actions, outputs, and risks. This explains why \texttt{compressed} can outperform \texttt{hint} on more technical tasks on SkillsBench despite using slightly more context.

\paragraph{All + Full as a diagnostic condition.}
\texttt{All + Full} performs worse than several compact alternatives, but we do not attribute this degradation to a single mechanism. It may reflect a combination of backend truncation or rejection, attention dilution, irrelevant procedural priors, and instruction conflict among many skills. Disentangling these factors would require dedicated controls over context length, relevance, and ordering. We therefore treat \texttt{All + Full} primarily as a diagnostic stress condition and place greater emphasis on the \texttt{Vector-top\(k\)} and \texttt{Graph-top\(k\)} comparisons, where candidate skills are
already filtered.

\begin{table*}[t]
\centering
\scriptsize
\resizebox{\linewidth}{!}{
\begin{tabular}{l|cc|cc|cc|cc|cc|cc}
\toprule
\multirow{3}{*}{\textbf{Configuration}}
& \multicolumn{6}{c|}{\textbf{ALFWorld}} 
& \multicolumn{6}{c}{\textbf{SkillsBench}} \\
\cmidrule(lr){2-7} \cmidrule(lr){8-13}
& \multicolumn{2}{c|}{\textbf{MiniMax}}
& \multicolumn{2}{c|}{\textbf{GPT-Codex}}
& \multicolumn{2}{c|}{\textbf{GLM}}
& \multicolumn{2}{c|}{\textbf{MiniMax}}
& \multicolumn{2}{c|}{\textbf{GPT-Codex}}
& \multicolumn{2}{c}{\textbf{GLM}} \\
& SR $\uparrow$ & C $\downarrow$
& SR $\uparrow$ & C $\downarrow$
& SR $\uparrow$ & C $\downarrow$
& AR $\uparrow$ & C $\downarrow$
& AR $\uparrow$ & C $\downarrow$
& AR $\uparrow$ & C $\downarrow$ \\
\midrule
Graph-top\(k\) + Full
& 66.4 & 4.4 & \underline{93.6} & 4.4 & \underline{80.0} & 4.4
& \underline{17.9} & 5.3 & \underline{32.8} & 5.3 & \underline{25.4} & 5.3 \\
Graph-top\(k\) + Hint
& \underline{68.6} & 0.07 & 92.1 & 0.07 & 76.4 & 0.07
& 16.3 & 0.12 & 30.5 & 0.12 & 23.9 & 0.12 \\
Graph-top\(k\) + Compressed
& \textbf{71.4} & 0.28 & \textbf{95.0} & 0.28 & \textbf{84.3} & 0.28
& \textbf{19.0} & 0.48 & \textbf{34.6} & 0.48 & \textbf{27.1} & 0.48 \\
Graph-top\(k\) + Workflow
& 63.6 & 1.4 & 91.4 & 1.4 & 75.7 & 1.4
& 17.4 & 1.7 & 31.5 & 1.7 & 24.6 & 1.7 \\
\bottomrule
\end{tabular}
}
\caption{
Structured-provider analysis under \texttt{skills\_1000}. \texttt{Graph-top\(k\)} uses Graph-of-Skills \citep{liu2026gos} as a representative graph-based provider. \(C\) denotes the token length of the rendered skill context, reported in thousands of tokens. The results show that interface choice remains important even with different skill selectors or providers. Task-level bootstrap 95\% confidence intervals
are reported in Appendix~\ref{app:uncertainty}.
}
\label{tab:provider_influence}
\end{table*}
\subsection{Cost--Performance Tradeoff}
\label{sec:cost_tradeoff}

A central motivation for SkillAlign is that exposure affects both utility and context cost. Figure~\ref{fig:cost_tradeoff} plots the MiniMax M2.7 cost--performance tradeoff on ALFWorld under the generic candidate sources in Table~\ref{tab:main_results}. We use this setting because its values directly correspond to one benchmark--backend block in the main table. Interfaces in the upper-left region achieve higher success with lower exposure cost.

Full exposure applies a strong but expensive procedural prior. Hint and compressed exposure apply weaker or more structured priors at much lower cost. When these compact interfaces achieve comparable or better success, exposing less skill content is not merely an efficiency trick; it is a better alignment between the agent's need and the skill's influence.

\subsection{Influence of Candidate Provider}
\label{sec:provider_influence}

The main experiments use generic candidate sources to avoid tying SkillAlign to a specific retrieval system. To test whether the exposure effect persists under structured skill selection, we additionally evaluate a graph-based candidate provider as shown in Table \ref{tab:provider_influence}. We refer to this setting as \texttt{Graph-top\(k\)}, instantiated with Graph-of-Skills \citep{liu2026gos}. This analysis checks whether the same exposure problem remains when candidate skills are selected by a structured provider.

The structured-provider results are consistent with the main findings. \texttt{Graph-top\(k\) + Compressed} is generally stronger than other graph-provider interfaces, while \texttt{Graph-top\(k\) + Full} is more costly and not consistently better. This suggests that better candidate selection does not eliminate the exposure problem. Even when a structured provider supplies a compact and relevant candidate set, the agent still benefits from receiving those skills through an aligned interface.

\subsection{Scaling with Skill-Library Size}
\label{sec:size_ablation}

\begin{figure}[t]
    \centering
    \includegraphics[width=\linewidth]{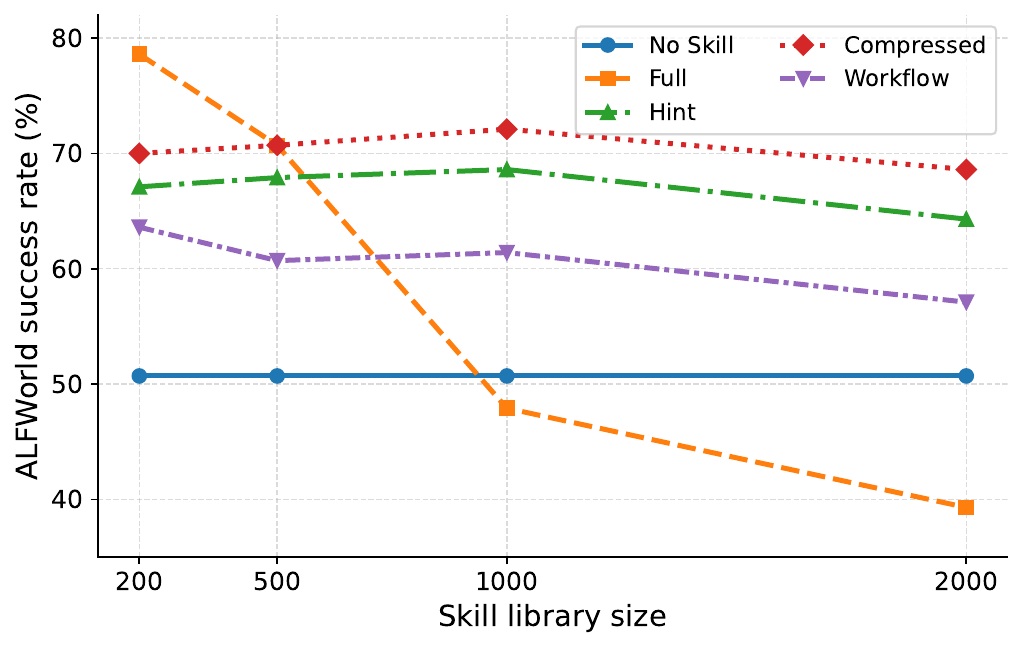}
    \caption{
    Reward trend under growing skill libraries. The no-skill baseline remains constant because no skill content is exposed. The 500-skill setting is slightly lower than the 1,000-skill setting for compact exposure, while 2,000 skills introduces more noise and cost.
    }
    \label{fig:size_reward}
\end{figure}

We evaluate whether exposure choice becomes more important as skill libraries grow. The evaluation task set is fixed, while the number of available skills varies from \texttt{skills\_200} to \texttt{skills\_500}, \texttt{skills\_1000}, and \texttt{skills\_2000}. We conduct this analysis on ALFWorld with MiniMax M2.7 under the all-candidate source, where library size directly controls the amount and diversity of available skill content. Figure~\ref{fig:size_reward} visualizes the reward trend across exposure interfaces.

The scaling results make the exposure problem more explicit. At smaller library sizes, \texttt{full} exposure can still be competitive because the raw skill context remains manageable. As the library grows, however, \texttt{full} exposure becomes increasingly expensive and degrades sharply, dropping to 47.9 at \texttt{skills\_1000}. Compact interfaces, especially \texttt{compressed}, remain more stable and reach 72.1 at \texttt{skills\_1000}. The 500-skill setting is slightly lower than the 1,000-skill setting for compact exposure, while performance drops again at 2,000 skills as the available skill pool becomes larger and noisier. These trends support the central motivation of SkillAlign: large skill ecosystems do not only require better retrieval, but also better control over how selected or available skills condition the agent.

\subsection{Exploratory Exposure Policy Learning}
\label{sec:policy_results}

We further test whether exposure choice can be learned from counterfactual supervision. The training data is generated from auxiliary ScienceWorld, where the same task/provider setting is evaluated under different fixed interfaces. We then evaluate the learned policy on ALFWorld through off-policy replay: the policy selects an interface for each task, and we look up the corresponding fixed-interface trajectory result. This experiment is intended as a feasibility probe rather than the core contribution.

\begin{figure}[t]
    \centering
    \includegraphics[width=\linewidth]{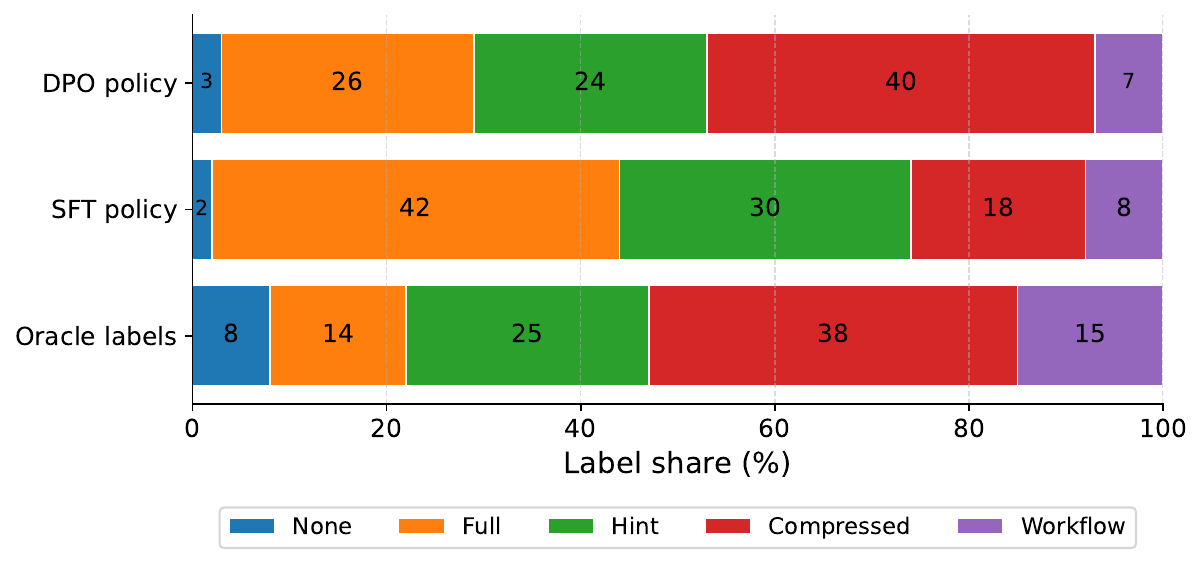}
    \caption{
    Distribution of exposure-interface labels. Oracle labels are diverse, indicating that no single exposure form is universally optimal. Learned policies partially recover this pattern but remain biased, explaining the remaining gap to oracle exposure.
    }
    \label{fig:policy_distribution}
\end{figure}

Table~\ref{tab:policy_learning} compares the best fixed interface, SFT policy, DPO policy, and oracle exposure under two candidate sources. Oracle exposure selects the best realized interface per task from fixed-mode counterfactuals and serves only as an analysis upper bound. Figure~\ref{fig:policy_distribution} compares the distribution of oracle labels and learned policy predictions.

\begin{table}[t]
\centering
\small
\setlength{\tabcolsep}{5pt}
\renewcommand{\arraystretch}{1.05}
\begin{tabular}{l|cc}
\toprule
\multirow{2}{*}{\textbf{Selector}}
& \multicolumn{2}{c}{\textbf{ALFWorld SR (\%)}} \\
& Vector-top\(k\) & Graph-top\(k\) \\
\midrule
Best Fixed Interface & 75.0 & 71.4 \\
\hline
SFT Policy           & 76.4 & 73.6 \\
+DPO Policy           & \underline{78.6} & \underline{76.4} \\
\hline
Oracle Exposure      & \textbf{92.9} & \textbf{85.7} \\
\bottomrule
\end{tabular}
\caption{
Exploratory exposure policy learning on ALFWorld. SFT and DPO policies are trained from auxiliary counterfactual outcomes and evaluated by replay. Oracle exposure estimates the headroom of perfect task-level interface selection and is not a deployable policy.
}
\label{tab:policy_learning}
\end{table}

Under replay evaluation, the learned policies improve over the best fixed interface, suggesting that exposure choice contains learnable signal, while also making clear that the current SFT/DPO setup is only a preliminary validation of feasibility. DPO improves over SFT in our replay evaluation, providing preliminary evidence that preference supervision can capture additional exposure-selection signal. However, the large oracle gap and the mismatch in label distributions indicate that more advanced policy learning remains an important direction for future work.

\subsection{Analysis}
\label{sec:analysis}
\paragraph{Fidelity of generated views.}
Since Hint, Compressed, and Workflow are generated from the raw skill, an observed failure could in principle arise from information omitted or distorted during card generation rather than from the exposure interface itself. We therefore audit 180 task--skill--view instances from 60 randomly sampled tasks (30 from ALFWorld and 30 from SkillsBench) using two independent human annotators. To avoid treating legitimate compression as an error, we count an omission only when information necessary for applying the skill to the paired task is absent. Table~\ref{tab:fidelity_audit} summarizes the results. Task-relevant omissions occur in 3.3\%, 5.0\%, and 6.7\% of the Hint, Compressed, and Workflow instances, respectively, while distortions, contradictions, and misleading additions are similarly rare.


\begin{table}[t]
\centering
\small
\begin{tabular}{lccc}
\toprule
\textbf{View} &
\textbf{Omission} &
\textbf{Dist./Contr.} &
\textbf{MA.} \\
\midrule
Hint & 2/60 (3.3\%) & 1/60 (1.7\%) & 1/60 (1.7\%) \\
Comp. & 3/60 (5.0\%) & 2/60 (3.3\%) & 1/60 (1.7\%) \\
Workflow   & 4/60 (6.7\%) & 2/60 (3.3\%) & 2/60 (3.3\%) \\
\bottomrule
\end{tabular}
\caption{Human fidelity audit of generated SkillAlign views. (Comp.: Compressed; Dist.: distortions; Contr.: contradictions; MA.: Misleading additions.)}
\label{tab:fidelity_audit}
\end{table}

Overall, the audit suggests that potential execution- or safety-relevant 
fidelity failures are uncommon under the current generation mechanism, 
but they are not completely eliminated. 
For safety-sensitive applications, compact views can additionally be 
paired with fidelity verification and fallback to the Full view when necessary.

\paragraph{Generic length-matched summarization.}
To separate exposure structure from context shortening, we further compare
Compressed against a generic free-form summary under both Vector-top-$k$
and Graph-top-$k$, across all three backends. Generic Summary uses the
same candidate skills, generation model, decoding settings, and
approximately matched rendered length as Compressed. Across the resulting
12 benchmark--provider--backend settings, Compressed generally outperforms
Generic Summary at a comparable context budget, while Full remains
competitive or stronger in some settings. This indicates that context
shortening explains part, but not all, of the benefit of compact exposure,
and further supports our conclusion that the appropriate interface is
setting-dependent. See full results in Appendix~\ref{app:generic}.

\paragraph{Candidate source and interface.}
Retrieval quality and exposure form are complementary. Top-\(k\) retrieval reduces irrelevant skill mass, while SkillAlign controls how the selected skills condition the agent. This explains why compact vector- or graph-based exposure can outperform both no-skill and full-library exposure.

\paragraph{Adaptive exposure.}
The oracle-policy gap shows that interface choice contains learnable signal, but our simple SFT/DPO policies only recover part of it. This suggests that adaptive exposure is promising, while stronger policy learning is needed to approach oracle-level interface selection.

\section{Conclusion}
\label{sec:conclusion}

We presented SkillAlign, a framework for studying and improving how skills are exposed to LLM agents. Rather than treating skills as static prompt snippets or assuming that more skill content is always better, SkillAlign views skills as externalized procedural priors whose effect depends on the interface through which they condition the agent. We instantiate this idea by reframing raw skills into multiple exposure interfaces, including full descriptions, hints, compressed summaries, and workflow-style guidance. 

Across ALFWorld and SkillsBench, our experiments show that exposure form is a first-class factor in skill-augmented agents. Full-library exposure can be harmful despite containing more information, while compact interfaces often achieve better cost--performance tradeoffs. We further show that exposure effects persist under different candidate providers, and that the best interface varies across tasks, backends, and skill sources. Finally, exploratory policy-learning results indicate that exposure choice is learnable, although current SFT/DPO policies remain far below oracle exposure. These findings suggest that future skill-agent systems should optimize not only skill acquisition and retrieval, but also the interface through which skills are presented to agents.

\section*{Limitations}
\label{sec:limitations}

Our work has several limitations. 

First, SkillAlign currently relies on a small set of manually designed exposure interfaces. Although these interfaces are sufficient to reveal the importance of skill exposure, they do not cover the full design space of possible skill representations. Future work could learn richer interfaces or generate task-adaptive views directly. 

Second, our exposure-policy learning is intentionally simple. The SFT and DPO policies are used mainly as feasibility probes, and the large gap to oracle exposure shows that adaptive exposure remains far from solved. Stronger policy models, better counterfactual data construction, and more robust preference filtering may further improve interface selection. 

Third, our framework is complementary to skill retrieval rather than a replacement for it. The quality of candidate skills still matters: poor retrieval can limit the benefit of even a well-aligned exposure interface. 
This suggests that skill-agent systems should jointly consider retrieval quality, exposure form, and downstream agent behavior.

\section*{Use of AI Assistants}
The authors used AI assistants (Claude Code and OpenAI Codex) for limited language editing, grammar checking, and code-debugging assistance. All research ideas, method design, experimental decisions, experiment execution, result analysis, and manuscript writing were conducted by the authors. The authors reviewed all AI-assisted outputs and take full responsibility for the final content.

\section*{Acknowledgements}
This work is supported by the Strategic Priority Research Program of Chinese Academy of Sciences under Grant XDA04080400 and Beijing Natural Science Foundation L259016.




\bibliography{custom}

\appendix

\section{Implementation and Evaluation Details}
\label{app:implementation}

\subsection{Provider-Agnostic Exposure Layer}

SkillAlign is implemented as a provider-agnostic exposure layer. Benchmarks and upstream providers supply task context, agent context, and candidate skills; SkillAlign decides how the candidate skills are rendered before they are inserted into the agent prompt. This separation keeps the exposure layer independent of benchmark-specific runners and retrieval systems. In particular, the core abstraction only assumes generic task, agent, and candidate-skill objects, rather than ALFWorld-, SkillsBench-, ScienceWorld-, or graph-retrieval-specific classes.

The output of SkillAlign is a rendered skill context together with metadata used for analysis. This metadata includes the selected skill ids, exposure interface, rendered length, and decision records. Since the same candidate skills can be rendered through different interfaces, the implementation naturally supports counterfactual comparisons where only exposure form changes.

\subsection{Candidate Providers}

SkillAlign separates candidate provision from exposure selection. In the experiments, we use three generic candidate sources and one structured provider for analysis (\(k=5\) for ALFWorld and \(k=8\) for SkillsBench):

\begin{itemize}
    \item \texttt{none}: no skill context is provided, serving as the no-exposure control.
    \item \texttt{all}: the full skill library is passed to SkillAlign, isolating exposure effects from retrieval.
    \item \texttt{vector-top\(k\)}: skills are retrieved through semantic similarity over cached skill representations. 
    \item \texttt{Graph-top\(k\)}: a representative graph-based provider used only to test whether exposure effects persist under structured candidate selection.
\end{itemize}

The provider identity is logged separately from the exposure interface. This allows us to distinguish failures caused by poor candidate selection from failures caused by inappropriate exposure.

\subsection{Benchmark Settings}

Table~\ref{tab:appendix_benchmark_settings} summarizes the main evaluation settings.

\begin{table}[t]
\centering
\small
\setlength{\tabcolsep}{4pt}
\renewcommand{\arraystretch}{1.12}
\begin{tabular}{p{0.26\linewidth}p{0.62\linewidth}}
\toprule
\textbf{Setting} & \textbf{Description} \\
\midrule
Main benchmarks & ALFWorld and SkillsBench. \\
Auxiliary data & ScienceWorld and earlier TextWorld runs are used for exposure-policy data construction. \\
Main skill library & \texttt{skills\_1000}. \\
Library-size analysis & \texttt{skills\_200}, \texttt{skills\_500}, \texttt{skills\_1000}, and \texttt{skills\_2000}. \\
ALFWorld episodes & 140 evaluation episodes with a maximum of 30 steps. \\
SkillsBench tasks & 94 containerized technical tasks. \\
Backends & \texttt{minimax-m2.7}, \texttt{gpt-5.3-codex}, and \texttt{glm-5}. \\
Interfaces & \texttt{none}, \texttt{full}, \texttt{hint}, \texttt{compressed}, and \texttt{workflow}. \\
\bottomrule
\end{tabular}
\caption{Summary of the main SkillAlign evaluation settings.}
\label{tab:appendix_benchmark_settings}
\end{table}

For each run, we log the task id, provider, selected skill ids, exposure interface, rendered context length, task reward or success, number of steps when available, backend model, and experiment metadata. These logs allow us to reconstruct cost--performance tradeoffs, interface distributions, negative-transfer cases, and oracle exposure results from fixed-interface trajectories.

\section{Skill Cards and Exposure Interfaces}
\label{app:skill_cards}

\subsection{Skill Card Schema}

A SkillAlign card stores multiple views of the same underlying skill. These views are designed for different cognitive roles, not merely for different lengths. Table~\ref{tab:skill_card_schema} summarizes the main fields.

\begin{table}[t]
\centering
\small
\setlength{\tabcolsep}{4pt}
\renewcommand{\arraystretch}{1.12}
\begin{tabular}{p{0.28\linewidth}p{0.62\linewidth}}
\toprule
\textbf{Field} & \textbf{Role} \\
\midrule
\texttt{skill\_id} & Unique skill identifier. \\
\texttt{name} & Human-readable skill name. \\
\texttt{description} & Short factual description. \\
\texttt{routing\_summary} & Use/avoid boundary for deciding whether the skill belongs in context. \\
\texttt{short\_hint} & One immediate action cue. \\
\texttt{compressed} & Compact operational specification. \\
\texttt{workflow} & Ordered procedure with checks and stopping conditions. \\
\texttt{raw\_text} & Original full skill document. \\
\bottomrule
\end{tabular}
\caption{SkillAlign card schema.}
\label{tab:skill_card_schema}
\end{table}

The \texttt{full} interface renders \texttt{raw\_text}; it is not an LLM-generated summary. This distinction is important because raw skills may contain execution-critical scripts, arguments, assumptions, examples, APIs, or configuration details. SkillAlign may summarize such information in the \texttt{compressed} or \texttt{workflow} views, but it does not modify the original skill artifact.

\subsection{Exposure Interface Roles}

Table~\ref{tab:appendix_interfaces} summarizes the five primary exposure interfaces.

\begin{table}[t]
\centering
\small
\setlength{\tabcolsep}{4pt}
\renewcommand{\arraystretch}{1.12}
\begin{tabular}{p{0.20\linewidth}p{0.68\linewidth}}
\toprule
\textbf{Interface} & \textbf{Role} \\
\midrule
\texttt{none} & Exposes no skill text. Used when candidate skills are irrelevant, misleading, or too costly. \\
\texttt{full} & Exposes the raw skill document. Useful when exact scripts, APIs, arguments, or implementation details are needed. \\
\texttt{hint} & Exposes one short action-oriented cue. Useful when the agent only needs a weak nudge. \\
\texttt{compressed} & Exposes trigger conditions, inputs/state, core action, expected output, and risks. \\
\texttt{workflow} & Exposes ordered steps with verification points and stopping conditions. Useful when sequencing matters. \\
\bottomrule
\end{tabular}
\caption{SkillAlign exposure interfaces and their intended roles.}
\label{tab:appendix_interfaces}
\end{table}

The legacy \texttt{checklist} field is retained only for backward compatibility and is not used as a primary experimental mode. Its checkpoint content is folded into the \texttt{workflow} view.

\subsection{Example Skill Card}

Table~\ref{tab:skill_card_example} gives a compact example for an ALFWorld inventory-management skill. The example illustrates how different views expose different information rather than simply shortening the same text.

\begin{table*}[t]
\centering
\small
\setlength{\tabcolsep}{5pt}
\renewcommand{\arraystretch}{1.12}
\begin{tabular}{p{0.18\linewidth}p{0.74\linewidth}}
\toprule
\textbf{Field} & \textbf{Example Content} \\
\midrule
\texttt{skill\_id} & \texttt{alfworld-inventory-management} \\
\texttt{routing\_summary} & Select for ALFWorld tasks where the agent must track carried objects, containers, receptacles, or object placement across rooms. Avoid for pure navigation when no object state must be remembered. \\
\texttt{short\_hint} & Check inventory after every pickup and before every put/open/heat/cool action. \\
\texttt{compressed} & Trigger: the task requires carrying, placing, heating, cooling, cleaning, or inspecting objects. Inputs/State: current observation, inventory, target object, receptacles, and appliance states. Do: maintain an explicit inventory and location note; verify object possession before acting on it; verify receptacle or appliance state before placing or transforming objects. Output: fewer invalid actions and clearer goal progress. Watch: using objects not in inventory, forgetting object locations, or acting before opening or activating containers. \\
\texttt{workflow} & Observe current room, visible target objects, candidate receptacles, and inventory. Before pickup, ensure the object is visible and reachable. Before put, heat, cool, clean, or examine actions, verify that the needed object is currently carried. After placing an object, remember the receptacle location and state. Stop when the environment gives success or the goal condition is visibly satisfied. \\
\bottomrule
\end{tabular}
\caption{Compact example of a SkillAlign card for an embodied-agent skill.}
\label{tab:skill_card_example}
\end{table*}

\begin{table*}[htp]
\centering
\scriptsize
\setlength{\tabcolsep}{3.2pt}
\resizebox{\textwidth}{!}{
\begin{tabular}{lllcccc}
\toprule
\textbf{Provider} &
\textbf{Benchmark} &
\textbf{Backend} &
\textbf{Full} &
\textbf{Hint} &
\textbf{Compressed} &
\textbf{Workflow} \\
\midrule

\multirow{6}{*}{All}
& ALFWorld & MiniMax M2.7
& 47.9 [39.8, 56.1]
& \underline{68.6 [60.7, 75.8]}
& \textbf{72.1 [64.2, 79.1]}
& 61.4 [53.2, 69.1] \\

& ALFWorld & GPT-Codex
& 88.6 [82.6, 93.3]
& \underline{92.1 [86.7, 95.8]}
& \textbf{93.6 [88.6, 96.8]}
& 90.0 [84.3, 94.2] \\

& ALFWorld & GLM-5
& 70.7 [62.7, 77.8]
& \underline{80.7 [73.5, 86.7]}
& \textbf{82.1 [75.0, 87.9]}
& 76.4 [69.0, 82.7] \\

& SkillsBench & MiniMax M2.7
& 14.5 [12.1, 17.1]
& \underline{16.8 [14.2, 19.6]}
& \textbf{17.2 [14.6, 20.0]}
& 15.6 [13.1, 18.2] \\

& SkillsBench & GPT-Codex
& 24.8 [21.6, 28.2]
& \underline{29.4 [25.9, 33.0]}
& \textbf{30.8 [27.2, 34.5]}
& 27.6 [24.2, 31.2] \\

& SkillsBench & GLM-5
& 20.4 [17.5, 23.5]
& \underline{24.0 [20.9, 27.3]}
& \textbf{25.1 [21.9, 28.4]}
& 22.8 [19.8, 26.0] \\

\midrule

\multirow{6}{*}{Vector-top-$k$}
& ALFWorld & MiniMax M2.7
& \underline{73.6 [66.0, 80.5]}
& \textbf{75.0 [67.4, 81.8]}
& 69.3 [61.3, 76.6]
& 64.3 [56.1, 72.0] \\

& ALFWorld & GPT-Codex
& 92.9 [87.4, 96.4]
& \textbf{94.3 [89.2, 97.4]}
& \underline{94.0 [88.8, 97.2]}
& 91.4 [85.5, 95.4] \\

& ALFWorld & GLM-5
& 80.7 [73.5, 86.7]
& \textbf{83.6 [76.7, 89.2]}
& \underline{82.9 [75.9, 88.6]}
& 78.6 [71.0, 84.9] \\

& SkillsBench & MiniMax M2.7
& \underline{16.4 [13.9, 19.1]}
& 15.6 [13.1, 18.2]
& \textbf{18.1 [15.4, 21.0]}
& 16.2 [13.7, 18.9] \\

& SkillsBench & GPT-Codex
& \underline{31.8 [28.2, 35.5]}
& 28.9 [25.4, 32.5]
& \textbf{32.2 [28.6, 35.9]}
& 29.6 [26.1, 33.2] \\

& SkillsBench & GLM-5
& \underline{24.8 [21.7, 28.1]}
& 23.4 [20.3, 26.6]
& \textbf{26.2 [22.9, 29.6]}
& 24.4 [21.2, 27.7] \\

\midrule

\multirow{6}{*}{Graph-top-$k$}
& ALFWorld & MiniMax M2.7
& 66.4 [58.3, 73.9]
& \underline{68.6 [60.7, 75.8]}
& \textbf{71.4 [63.4, 78.6]}
& 63.6 [55.4, 71.3] \\

& ALFWorld & GPT-Codex
& \underline{93.6 [88.6, 96.8]}
& 92.1 [86.7, 95.8]
& \textbf{95.0 [90.2, 97.8]}
& 91.4 [85.9, 95.3] \\

& ALFWorld & GLM-5
& \underline{80.0 [72.7, 86.1]}
& 76.4 [69.0, 82.7]
& \textbf{84.3 [77.5, 89.7]}
& 75.7 [68.2, 82.1] \\

& SkillsBench & MiniMax M2.7
& \underline{17.9 [15.2, 20.8]}
& 16.3 [13.8, 19.0]
& \textbf{19.0 [16.2, 22.0]}
& 17.4 [14.8, 20.2] \\

& SkillsBench & GPT-Codex
& \underline{32.8 [29.2, 36.5]}
& 30.5 [26.9, 34.2]
& \textbf{34.6 [30.9, 38.4]}
& 31.5 [27.9, 35.2] \\

& SkillsBench & GLM-5
& \underline{25.4 [22.2, 28.8]}
& 23.9 [20.8, 27.2]
& \textbf{27.1 [23.8, 30.6]}
& 24.6 [21.4, 27.9] \\

\bottomrule
\end{tabular}
}
\caption{
Task-level bootstrap 95\% confidence intervals for SkillAlign under
the three evaluated candidate providers. Intervals are obtained from
10,000 bootstrap resamples of the evaluation tasks. Bold and underlined
values denote the highest and second-highest point estimates,
respectively, within each provider--benchmark--backend setting and do
not imply statistically significant rankings. The intervals quantify
uncertainty under task resampling rather than backend run-to-run
stochasticity.
}
\label{tab:bootstrap_ci_full}
\end{table*}

\section{Supplementary Analysis}
\subsection{Uncertainty Analysis}
\label{app:uncertainty}
We estimate uncertainty for all performance results in
Tables~\ref{tab:main_results} and~\ref{tab:provider_influence} using
task-level bootstrap resampling. For each configuration, we resample
the evaluation tasks with replacement 10,000 times and recompute the
success rate (ALFWorld) or average reward (SkillsBench). We report
percentile-based 95\% confidence intervals in Table \ref{tab:bootstrap_ci_full}. These intervals quantify
uncertainty across evaluation tasks rather than backend run-to-run
stochasticity.

Overall, task resampling preserves the main qualitative patterns of the
point estimates, while closely spaced results often have substantially
overlapping confidence intervals. Such small numerical differences
should therefore not be over-interpreted. This is consistent with our
main conclusion that no single exposure interface is universally
superior; instead, the effect of exposure depends on the task, backend,
and candidate provider.

\subsection{Generic Length-matched Summarization}
\label{app:generic}

To distinguish the effect of exposure structure from context shortening,
we compare the structured Compressed interface against a generic free-form summarization baseline in Table \ref{tab:generic_summary_full}. We evaluate this baseline on the full
ALFWorld and SkillsBench evaluation sets under both \texttt{Vector-top\(k\)} and
\texttt{Graph-top\(k\)}, using all three agent backends. For each setting,
Full, Generic Summary, and Compressed use the same
candidate skill ids and ordering. Generic Summary uses the same generation
model and decoding settings as Compressed, with approximately
matched rendered context length.

\begin{table*}[htp]
\centering
\small
\setlength{\tabcolsep}{5pt}
\begin{tabular}{lllccc}
\toprule
\textbf{Provider} &
\textbf{Benchmark} &
\textbf{Backend} &
\textbf{Full} &
\textbf{Generic Summary} &
\textbf{Compressed} \\
\midrule

\multirow{6}{*}{Vector-top$k$}
& ALFWorld
& MiniMax M2.7
& \textbf{73.6}
& 67.9
& \underline{69.3} \\

& ALFWorld
& GPT-Codex
& 92.9
& \underline{93.4}
& \textbf{94.0} \\

& ALFWorld
& GLM-5
& 80.7
& \underline{81.8}
& \textbf{82.9} \\

& SkillsBench
& MiniMax M2.7
& 16.4
& \underline{17.2}
& \textbf{18.1} \\

& SkillsBench
& GPT-Codex
& 31.8
& \underline{31.9}
& \textbf{32.2} \\

& SkillsBench
& GLM-5
& 24.8
& \underline{25.4}
& \textbf{26.2} \\

\midrule

\multirow{6}{*}{Graph-top$k$}
& ALFWorld
& MiniMax M2.7
& 66.4
& \underline{68.7}
& \textbf{71.4} \\

& ALFWorld
& GPT-Codex
& 93.6
& \underline{94.2}
& \textbf{95.0} \\

& ALFWorld
& GLM-5
& 80.0
& \underline{82.1}
& \textbf{84.3} \\

& SkillsBench
& MiniMax M2.7
& 17.9
& \underline{18.4}
& \textbf{19.0} \\

& SkillsBench
& GPT-Codex
& 32.8
& \underline{33.6}
& \textbf{34.6} \\

& SkillsBench
& GLM-5
& 25.4
& \underline{26.1}
& \textbf{27.1} \\

\bottomrule
\end{tabular}
\caption{
Generic length-matched summarization analysis on the full evaluation sets.
\texttt{Vector-top\(k\)} uses $k=5$ for ALFWorld and $k=8$ for SkillsBench.
Generic Summary uses the same candidate skills and approximately the same
rendered context budget as Compressed.
Bold and underlined values denote the highest and second-highest point
estimates within each row, respectively.
}
\label{tab:generic_summary_full}
\end{table*}

At comparable compact context lengths, Compressed generally
outperforms Generic Summary across providers, backends, and benchmarks,
while Full remains stronger in some settings. These results
suggest that context shortening explains part, but not all, of the benefit
of compact exposure. More broadly, they are consistent with our central
finding that the appropriate exposure interface is setting-dependent,
rather than universally fixed.

\section{Prompt Templates}
\label{app:prompts}

\subsection{Skill Card Generation}

The LLM-backed card generator is instructed to produce faithful structured views of the source skill. The prompt emphasizes that the views should play different cognitive roles and that execution contracts should be preserved when present.

\begin{promptbox}
You compile SkillAlign skill cards for LLM agents. Return only valid JSON.
Be faithful to the source and do not invent unavailable tools, credentials, APIs, or guarantees.
Create interface views with different cognitive functions, not merely different lengths.
Treat checklist as legacy checkpoint material; the main procedural interface is workflow.
Prefer precise operational language over generic advice.
When script/config evidence is provided, preserve the execution contract.
\end{promptbox}

The card-generation input contains the skill id, name, heuristic views, required output schema, interface distinctions, available evidence, and compacted source text. The generator produces fields such as \texttt{routing\_summary}, \texttt{short\_hint}, \texttt{compressed}, \texttt{workflow}, \texttt{preconditions}, \texttt{failure\_modes}, and \texttt{keywords}. The quality rules require the model to preserve scripts, arguments, dependencies, inputs, outputs, validation commands, state checks, side effects, and stopping conditions when they appear in the source.

\subsection{Agent-Facing Injection}

All exposed skill contexts begin with a shared wrapper that instructs the agent to treat skills as optional support rather than hard commands.

\begin{promptbox}
SkillAlign guidance:
Use the following skill interfaces only when they directly help the next action.
Treat them as optional support, not as a command to ignore the current observation.
\end{promptbox}

The interface-specific body is then appended. For example, the \texttt{workflow} interface uses ordered steps:

\begin{promptbox}
=== Skill Workflow: <skill name> ===
<routing summary>
1. <step with check>
2. <step with dependency>
3. <step with stop condition>
\end{promptbox}

The \texttt{compressed} interface uses an operational specification:

\begin{promptbox}
=== Compressed Skill: <skill name> ===
Trigger: ...
Inputs/State: ...
Do: ...
Output: ...
Watch: ...
\end{promptbox}

The \texttt{hint} interface contains only a short cue, while the \texttt{full} interface preserves the raw skill document.

\subsection{Learned Exposure Policy Prompt}

The learned exposure policy predicts an interface, not a skill set. Candidate skills have already been supplied by the provider. A compact prompt skeleton is shown below.

\begin{promptbox}
Choose the best SkillAlign exposure interface for this task and selected candidate skills.
Return only JSON with key interface_mode.
Do not choose which skills to retrieve. The provider has already selected the candidates.

Task ID: <task id>
Agent model: <backend model>
Available modes: none, full, hint, compressed, workflow

Task:
<task description>

Initial observation:
<optional observation>

Candidate skill count: <k>
Candidate skill ids: <ids>

Selected candidate skill cards:
<routing summary>
<hint view>
<compressed view>
<workflow view>
<full raw text status>

Decision criterion:
Choose the mode that maximizes expected task success while avoiding unnecessary context cost and negative transfer.

Return only:
{"interface_mode": "<mode>"}
\end{promptbox}

\section{Exposure Policy Data and Replay Evaluation}
\label{app:policy_training}

\subsection{Counterfactual Supervision}
We construct exposure-policy data from auxiliary counterfactual runs. For a task and candidate provider, the same candidate skills are evaluated under multiple fixed exposure interfaces. The resulting reward, rendered context cost, and trajectory length are used to derive oracle-style labels and pairwise preferences.

Given a task \(x\) and interface \(m\), we compute a utility score:

\[
U_x(m)=R_x(m)-\alpha C_x(m)-\beta T_x(m),
\]

where \(R_x(m)\) is task reward or success, \(C_x(m)\) is normalized rendered context cost, and \(T_x(m)\) is normalized trajectory length. The highest-utility interface provides an SFT label, while interface pairs with sufficient utility margin provide DPO preference pairs.

\subsection{Policy Learning}

The policy model receives the task description, provider metadata, interface definitions, and selected skill-card content, and predicts a single interface label for the candidate set. We use a single label per task-provider pair because the fixed-mode counterfactuals observe outcomes at the candidate-set level. This also keeps the policy aligned with the main experimental question: how should the selected skill set be exposed to the agent?

SFT provides a warm start from oracle-style labels, while DPO uses pairwise counterfactual preferences to distinguish cases where one interface is more useful or less costly than another. In this study, we do not use online reinforcement learning because it would require substantially more environment interaction and would confound exposure-policy learning with additional exploration.

\subsection{Replay Evaluation}

For learned-policy evaluation on ALFWorld, we use off-policy replay. The learned policy predicts an interface for each task and provider, and the evaluator looks up the corresponding fixed-interface trajectory result. This is much cheaper than re-running every environment episode, but it is reported as replay evaluation rather than fresh online interaction.

The learned-policy results should therefore be interpreted as a policy-learning analysis rather than as a fully optimized deployment setting. We report both reward and interface-selection distribution, and compare learned policies against oracle exposure to estimate the remaining headroom for adaptive interface selection.

\subsection{Training Details}

The learned exposure policy is trained separately from the main agent backends. It is used only for the replay-based exposure-selection analysis in Section~\ref{sec:policy_results}. We train LoRA adapters on Qwen3-4B model. The policy receives the task description, provider metadata, candidate skill-card content, and interface definitions, and predicts one exposure interface for the selected candidate skill set. The training parameters are shown in Table \ref{tab:policy_training_details}.

\begin{table}[t]
\centering
\small
\setlength{\tabcolsep}{4pt}
\renewcommand{\arraystretch}{1.12}
\begin{tabular}{p{0.30\linewidth}p{0.58\linewidth}}
\toprule
\textbf{Item} & \textbf{Setting} \\
\midrule
SFT data & Card-aware oracle-style interface labels from counterfactual outcomes \\
DPO data & Pairwise preferences between exposure interfaces under the same task/provider setting \\
SFT epochs & 5 \\
DPO epochs & 2 \\
SFT learning rate & \(2\times10^{-4}\) \\
DPO learning rate & \(5\times10^{-6}\) \\
DPO preference coefficient & \(\beta=0.1\) \\
Per-device batch size & 1 \\
Gradient accumulation steps & 16 \\
Precision & bfloat16 \\
\bottomrule
\end{tabular}
\caption{Training settings for the learned exposure-policy model.}
\label{tab:policy_training_details}
\end{table}

\section{Computational Budget}
The main evaluation cost comes from running fixed-interface counterfactual experiments under multiple candidate providers, exposure interfaces, benchmarks, and agent backends. ALFWorld evaluations are lightweight text-environment runs, while SkillsBench evaluations are more expensive because they require containerized task execution. The learned exposure-policy model is trained with LoRA and is substantially cheaper than full model training. Our training takes 1 NVIDIA A100 GPU for approximately 50 GPU hours in total (including trial experiments).

In our implementation, vector retrieval uses cached embeddings and local NumPy matrix search rather than a separate vector database. SkillsBench experiments require containerized execution, so disk usage and image startup failures are monitored during evaluation. 

\end{document}